# Hyperspectral Image Recovery Constrained by Multi-Granularity Non-Local Self-Similarity Priors

Zhuoran Peng, Yiqing Shen *IEEE*

*Abstract*—Hyperspectral image (HSI) recovery, as an upstream image processing task, holds significant importance for downstream tasks such as classification, segmentation, and detection. In recent years, HSI recovery methods based on non-local prior representations have demonstrated outstanding performance. However, these methods employ a fixed-format factor to represent the non-local self-similarity tensor groups, making them unable to adapt to diverse missing scenarios. To address this issue, we introduce the concept of granularity in tensor decomposition for the first time and propose an HSI recovery model constrained by multi-granularity non-local self-similarity priors. Specifically, the proposed model alternately performs coarse-grained decomposition and fine-grained decomposition on the non-local self-similarity tensor groups. Among them, the coarse-grained decomposition builds upon Tucker tensor decomposition, which extracts global structural information of the image by performing singular value shrinkage on the mode-unfolded matrices. The fine-grained decomposition employs the FCTN decomposition, capturing local detail information through modeling pairwise correlations among factor tensors. This architectural approach achieves a unified representation of global, local, and non-local priors for HSIs. Experimental results demonstrate that the model has strong applicability and exhibits outstanding recovery effects in various types of missing scenes such as pixels and stripes.

## I. INTRODUCTION

As a cutting-edge remote sensing technique, hyperspectral imaging enables many fields to achieve a qualitative leap in the depth of information acquisition and analysis. Hyperspectral image (HSI) can simultaneously capture detailed information about target objects in both the spatial and spectral dimensions, tightly integrating the traditional two-dimensional spatial information with spectral information to form a three-dimensional data cube. This enables people to more comprehensively and accurately identify and distinguish different material components, land cover types, and so on. In numerous fields such as agricultural monitoring [1], military reconnaissance [2], geological exploration [3], and medical diagnosis [4], HSI is playing an increasingly critical role. Its research holds extremely significant importance for advancing technological progress in related fields and enhancing the scientific rigor and accuracy of decision-making.

However, due to various reasons, HSI may experience information loss. During the information acquisition process, factors such as sensor performance limitations, the complexity of the imaging environment, and data transmission interference can easily lead to information loss or distortion in HSI [5], [6], [7]. Sensor noise interference can cause errors in spectral information, affecting material composition identification; environmental factors such as atmospheric scattering and uneven illumination can reduce image contrast and blur details, which is detrimental to target detection and classification; signal attenuation or malfunctions during data transmission can also lead to the loss of some data. This loss of information severely impacts subsequent applications of HSI, such as classification and detection [8], [9], [10]. Therefore, conducting research on HSI recovery, through effective algorithms and technologies to recover lost information, enhance image quality and usability, is of great significance for leveraging its advantages and meeting the demands for high-precision images.

In recent years, HSI recovery based on prior representations has garnered significant attention from numerous researchers [11], [12]. Prior representations, as the name suggests, refer to the use of known prior knowledge about HSI during the recovery process to guide the design and implementation of recovery algorithms. These prior pieces of knowledge can come from an understanding of the physical properties of HSI, prior knowledge of the imaging scene, and analysis of the statistical properties of the image, among others. For example, HSI typically exhibits spectral smoothness priors [13], [14], meaning that pixel values change gradually along the spectral dimension. This property can be used as a constraint to help recover missing spectral information. Additionally, the spatial non-local self-similarity prior of images [15], [16] indicates that even image patches that are spatially far apart can exhibit similar features. This is helpful for repairing image distortions caused by noise or interference. By incorporating these prior pieces of knowledge into the recovery model, it is possible to more effectively suppress noise and fill in missing data, thereby significantly improving the quality and reliability of the restored images. This provides a more accurate data foundation for subsequent HSI analysis and applications.

¹This work was supported by Toursun Synbio Inc., Shanghai 201210, China. (Corresponding author: Yiqing Shen.)
Zhuoran Peng is with the Courant Institute of Mathematical Sciences, New York University, New York, NY 10012 USA (e-mail: zp2019@nyu.edu).
Yiqing Shen is with Toursun Synthetic Biology Inc., Shanghai 201210, China (e-mail: yiqingshen1@gmail.com).

As a natural representation form for HSI data [17], [18], tensors excel in exploiting the correlations within multidimensional data. The integration of tensor models with prior knowledge provides strong support for HSI recovery. By representing HSI as a three-dimensional tensor and utilizing tensor decomposition techniques, it is possible to uncover its low-rank characteristics. Combining this with prior knowledge such as spectral smoothness and spatial non-local self-similarity allows for more effective recovery of lost information, noise suppression, and improved image quality. This fusion approach can also be extended to multi-source data fusion scenarios, further enriching the information and providing robust support for complex scene analysis.

Tensor decomposition and tensor networks are important tools for processing high-dimensional data. Tucker decomposition [19] achieves this by decomposing a tensor into a core tensor and a set of factor matrices for each mode. This approach preserves the multi-dimensional structure and correlations within the data, making it suitable for compressing and analyzing complex datasets. CP (CANDECOMP/ARAFAC) decomposition [20] represents tensors as the sum of multiple rank-1 tensors, which has a concise expression and is suitable for sparse data and fast computation. T-SVD (Tensor Singular Value Decomposition) [21] is an extension of matrix SVD to tensors. By decomposing the tensor after Fourier transformation, it can effectively extract the main components of the data, making it suitable for noise removal and feature extraction. In the realm of tensor networks, the TT (Tensor Train) [22] format significantly reduces data storage and computational complexity by decomposing high-dimensional tensors into a chain of lower-dimensional tensors, making it suitable for large-scale data processing. The TR (Tensor Ring) [23] format further optimizes tensor representation by employing a ring-like decomposition structure, enabling more efficient handling of data with cyclic correlations. The FCTN (Fully Connected Tensor Network) [24] introduces fast transform techniques to accelerate the computation process within tensor networks, thereby improving processing efficiency. These methods can effectively uncover the intrinsic structures within HSI, enhancing both the efficiency and accuracy of data processing.

Applying tensor decomposition to the entire HSI can effectively exploit global priors, but it overlooks the inherent non-local self-similarity (NSS) characteristics of HSI. To address this issue, some researchers have sought non-local self-similar group tensors within the HSI and then further decomposed these tensors to exploit non-local priors. Song et al. [25] used T-SVD to decompose non local self similar tensor groups and conducted error analysis of the algorithm to effectively mine global and non local priors. Xie et al. [26] performed singular value shrinkage on the mode unfolded matrices of non-local self-similar group tensors to effectively exploit their low Tucker rank, and experiments demonstrated the outstanding performance of the algorithm. Zhang et al. [27] used triangulation-based linear interpolation to initialize the image, then grouped similar non-local blocks into a tensor for completion. Additionally, they proposed block matching and block mismatching mechanisms, providing a novel approach for tensor-based non-local methods.

Due to the fact that tensor networks can further exploit the details of HSI through more complex factor connections, the integration of tensor networks with non-local priors has become a hot topic. Ding et al. [28] integrated the TT network into a non-local framework, combined with ket augmentation technology, resulting in a significant improvement in recovery performance. Chen et al. [29] proposed a non-local tensor ring decomposition method and applied it to the HSI denoising task. Experiments demonstrated that its noise removal performance outperformed that of pure tensor methods. Zheng et al. [30] fused the FCTN network with a non-local block method. The introduction of a high-order block matching mechanism brought about a qualitative leap in performance.

Although the aforementioned methods have achieved significant breakthroughs in HSI recovery, several limitations remain: Traditional tensor decomposition methods (e.g., Tucker decomposition) primarily focus on dimensionality reduction by mapping HSI principal components to tensor factors. These methods typically produce large-dimensional tensor factors with coarse-grained characteristics. While effective for extracting global structural information, they demonstrate limited capability in capturing fine details; Tensor network methods (e.g., FCTN) establish complex factor interactions through small-dimensional tensor factors with fine-grained characteristics. Although capable of extracting detailed information through factor interactions, their fine-grained nature may compromise the original HSI structure, consequently weakening their ability to represent global structural information.

To address these limitations, this paper proposes a multi-granularity non-local self-similarity constrained HSI recovery method that unifies conventional tensor decomposition and emerging tensor networks within a non-local framework, capable of simultaneously capturing global structural information, local detailed features, and non-local self-similarity characteristics of HSI, with the main contributions summarized as follows:

(1) We propose a Multi-Granularity Non-local Self-Similarity (MG-NSS) prior for HSI recovery, which innovatively integrates both coarse-grained and fine-grained modules within the conventional non-local self-similarity framework. The coarse-grained module employs Tucker tensor decomposition to capture global structural information, while the fine-grained module utilizes FCTN decomposition to extract local detailed features. This novel architecture effectively unifies global, local, and non-local priors for comprehensive HSI representation.

(2) An efficient optimization algorithm is developed based

on ADMM and PAM frameworks, where the coarse-grained and fine-grained modules alternately identify similar tensor patches and perform completion operations. Extensive experimental results demonstrate the outstanding performance of our method in HSI recovery tasks, achieving excellent reconstruction quality even under extremely low sampling rates (SR=0.01) and showing superior advantages compared to state-of-the-art baseline methods.

The overall content structure of this paper is organized as follows: This section mainly introduces the HSI recovery task and some related work, and introduces the innovations of this paper; Section 2 is the preliminary knowledge section, which will introduce the Tucker decomposition and FCTN decomposition used in this paper; Section 3 is the method section, which mainly introduces the overall framework and various modules of the MG-NSS model proposed in this paper; Section 4 is the experiment, which records the results on multiple datasets to demonstrate the effectiveness and performance improvement of the proposed method; Finally, Section 5 summarizes this paper.

II. PRELIMINARIES

Based on references [31] and [24], this section introduces the tensor completion model based on Tucker decomposition and the tensor completion model based on FCTN decomposition, providing a foundation for the method proposed in Section 3.

*A. Tensor Completion Model Based on Tucker Decomposition*

Liu et al. [31] introduced Tucker decomposition into the tensor completion model and developed several solution algorithms. They formulated the tensor completion model based on Tucker decomposition as a problem of minimizing the nuclear norm of the tensor unfolding matrix under linear constraints:

$$\min_{\mathcal{X}} \sum_{k=1}^{N} \alpha_i \|\mathcal{X}_{(k)}\|_*$$
$$s.t. \ \mathcal{X}_\Omega = \mathcal{T}_\Omega \qquad (1)$$

where $\mathcal{X}$ represents the completed tensor, $\mathcal{T}$ denotes the observed tensor, $\mathcal{X}_{(k)}$ denotes the mode-k unfolding matrix of tensor $\mathcal{X}$, $\|\cdot\|_*$ represents the nuclear norm, and $\cdot_\Omega$ is a projection operation used to constrain the entries of tensors $\mathcal{X}$ and $\mathcal{T}$ to be equal within the set $\Omega$.

*B. Tensor Completion Model Based on FCTN Decomposition*

Zheng et al. [24] proposed a novel tensor network decomposition called FCTN and applied it to the task of tensor completion. The constructed model is as follows:

$$\min_{\mathcal{X},\mathcal{G}} \frac{1}{2} \|\mathcal{X} - \text{FCTN}(\mathcal{G}_1,\mathcal{G}_2,\cdots,\mathcal{G}_N)\|_F^2$$
$$s.t. \ \mathcal{X}_\Omega = \mathcal{T}_\Omega \qquad (2)$$

where $\mathcal{G}_i$ represents the factor tensor of FCTN, and $\|\cdot\|_F$ represents the Frobenius norm of the tensor.

III. METHOD

In this section, we first introduce the overall architecture of the proposed MG-NSS model. Then, we provide detailed explanations of its initialization and non-local iterative processes separately. Finally, we outline the algorithmic process through pseudo-code.

*A. Overall Architecture of MG-NSS Model*

Fig. 1 illustrates the architecture of the proposed MG-NSS model. The input HSI is first subjected to initial completion via Tucker decomposition and FCTN decomposition, yielding an initially completed HSI. This initially completed HSI is then fed into the main completion part of the model, which primarily consists of a coarse-grained completion module and a fine-grained completion module. The output of the coarse-grained completion module serves as the input to the fine-grained completion module, and then the output of the fine-grained completion module is used again as the input to the coarse-grained completion module. This process alternates iteratively to refine the HSI completion.

**Definition 1 (Granularity of Tensor Decomposition):** The granularity of tensor decomposition refers to the level of detail in partitioning or decomposing the original data during the tensor decomposition process, specifically reflected in the dimensions of the resulting factor tensors. It is categorized into coarse-grained and fine-grained based on the size. We define coarse-grained decomposition as methods that break down data into factors with larger dimensions, such as CP decomposition and Tucker decomposition. This type of decomposition focuses on global dimensionality reduction of the entire tensor without delving into specific details along each dimension. Conversely, we define fine-grained decomposition as methods that break down data into factors with smaller dimensions, such as TT, TR, and FCTN tensor network decomposition methods. This granularity focuses on processing details.

Based on this, we propose an HSI recovery model based on multi-grained tensor decomposition. This model integrates both coarse-grained and fine-grained tensor decompositions, enabling it to explore the global structure and local details of HSI.

*B. Initialization*

In the model's initialization module, let the input HSI be $\mathcal{T}$, the set of observed entries be $\Omega$, and the restored HSI be $\mathcal{X}$. The initialization process undergoes coarse-grained initialization followed by fine-grained initialization, and then the final initialization result is input into the non-local iterative optimization module.

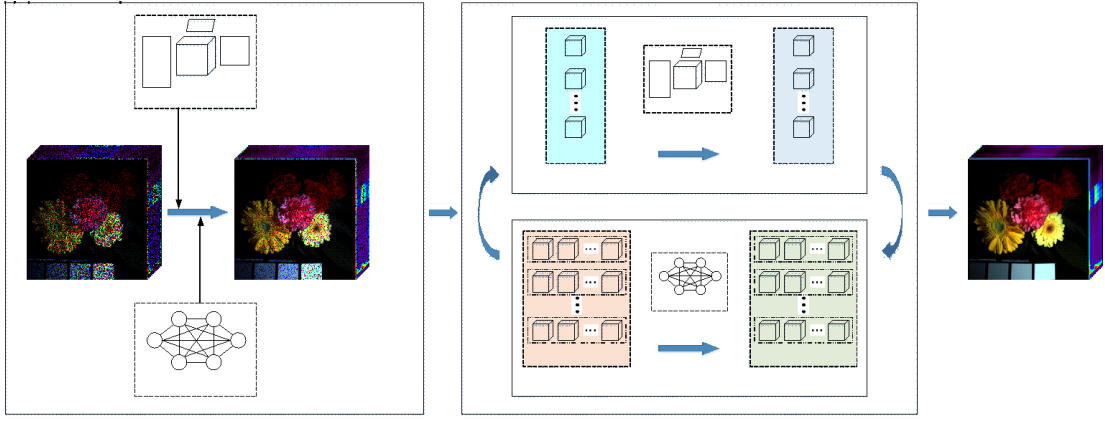

**Fig. 1.** Architecture of the MG-NSS model.

**1) Coarse-Grained Initialization**

Coarse-grained initialization refers to the completion of the entire HSI based on Tucker decomposition as follows:

$$\min_{X} \sum_{k=1}^{N} \alpha_i R_\varepsilon(X_{(k)})$$
$$s.t. \ X_\Omega = T_\Omega \quad (3)$$

here, $R_\varepsilon(X_{(k)}) = \sum_i \log(\sigma_i(X_{(k)}) + \varepsilon)$, which is used to constrain the low-rankness of the mode unfolded matrices. This model achieves the representation of the HSI's low-rank prior by minimizing the rank of the mode unfolded matrices based on Tucker decomposition. It performs initial completion of the HSI at the coarse-grained level. According to the literature [26], this coarse-grained initial completion can be achieved using the ADMM algorithm as follows:

$$\begin{cases} M_k^{t+1} = Fold_k\left(SVT\left(X_{(k)}^t + (1/\mu^t)\Lambda_k^t\right)\right) \\ X^{t+1} = \left((1/N)\sum_{k=1}^{N}\left(M_k^{t+1} - (1/\mu^t)\Lambda_k^t\right)\right)_{\Omega^C} + T_\Omega \\ \Lambda_k^{t+1} = \Lambda_k^t - \mu^t\left(M_k^{t+1} - X^{t+1}\right) \\ \mu^{t+1} = \eta * \mu^t \end{cases}$$
(4)

among them, $M$ is the auxiliary tensor of $X$, $\Lambda$ is the Lagrange multiplier, $\mu$ is a parameter greater than 0, $\eta$ is the step size, and $t$ is the number of iterations in the coarse-grained initialization.

**2) Fine-Grained Initialization**

Let the initial completed HSI obtained from coarse-grained initialization be $X_{CI}$, and let $X = X_{CN}$, yielding the following fine-grained initialization based on FCTN:

$$\min_{X,G} \frac{1}{2}\|X - FCTN(G_1, G_2, \cdots, G_N)\|_F^2$$
$$s.t. \ X_\Omega = T_\Omega \quad (5)$$

here, $FCTN(\cdot)$ represents the fully connected tensor network decomposition, and $G$ is its tensor network factor. This model decomposes the HSI $X$ using FCTN. Due to the small size of factor $G$, the features of the HSI are mapped into several low-dimensional spaces, achieving an initial completion of the HSI at the fine-grained level. According to the literature [30], this fine-grained initial completion can be solved based on the PAM algorithm as follows:

$$\begin{cases} G_i^{p+1} = GenFold\begin{pmatrix} [X_{(i)}^p(M_i^p)_{[n_1:N-1;m_1:N-1]} + \rho(G_i^p)_{(i)}] \\ *[(M_i^p)_{[m_1:N-1;n_1:N-1]}(M_o^p)_{[n_1:N-1;m_1:N-1]} + \rho I]^{-1} \end{pmatrix} \\ X^{p+1} = \left(\dfrac{FCTN\left(\{G_i^{p+1}\}_{i=1}^N\right) + \rho X^p}{1+\rho}\right)_{\Omega^C} + T_\Omega \end{cases}$$
(6)

where $GenFold(\cdot)$ is the inverse operation of generalized tensor unfolding, $X_{(i)} = (G_i)_{(i)}(M_i)_{[m_1:N-1;n_1:N-1]}$, $M_i$ is the combination of all $G$ except $G_i$, when $j < i$, $m_j = 2j, n_j = 2j-1$, when $j \geq i$, $m_j = 2j-1, n_j = 2j$. The output $X_{FI}$ after multiple iterations of formula (6) is taken as the final initialization result.

*C. Non-local Iterative Optimization*

After the initially completed HSI enters the non-local iterative optimization part of the model, it alternately performs coarse-grained and fine-grained completion on the NSS group of tensors.

The coarse-grained completion module identifies several groups of similar image patches within the HSI, then constructs these groups of similar image patches into a third-order tensor. By performing a completion algorithm based on Tucker decomposition on these tensors, it achieves the exploration of global structural information in the HSI. The fine-grained completion module also identifies several groups of similar image patches within the HSI, then constructs these groups of similar image patches into a fourth-order tensor. By performing a completion algorithm based on FCTN decomposition on these tensors, it achieves the exploration of local detail information in the HSI.

In HSI recovery models based on NSS prior representation, a critical step is the method for finding non-local self-similar blocks. To enhance the ability of the coarse-grained non-local module and the fine-grained non-local module proposed in this paper to capture the global structural information and local detail information of the HSI, respectively, we have analyzed this step.

The selection of non-local self-similar blocks currently widely used mainly includes K-means++ and block matching. The reason why methods based on K-means++ are more suitable for capturing image structural information is that by clustering image patches into different categories, they essentially seek blocks that are similar in overall patterns, even if they differ in local details. This clustering process focuses more on the "commonalities" or "macro-features" between patches, such as similar edge directions, general shape outlines, or the overall distribution of textures, thereby effectively extracting the skeleton structure of the image. In contrast, methods based on similar block matching directly calculate pixel-level or feature-level similarities between image patches to find matches, focusing more on the "micro-features" or "exact correspondences" between patches. This approach is like searching for almost identical "small fragments" within the image, making it more capable of capturing fine textures, specific local patterns, or subtle structural changes (i.e., detail information) that recur in the image.

Based on the above analysis, our coarse-grained non-local module and fine-grained non-local module employ K-means++ and block matching methods respectively to identify similar image patches, enabling efficient representation of both structural and detailed HSI characteristics. The implementation processes of these two modules will be elaborated in detail below.

**1) Coarse-Grained Non-Local Module**

Let the input HSI of this module be $\mathcal{X} = \mathcal{X}^{[f,u]}$, where $f$ is a flag, $u$ is the round of non-local iterations, indicating that the input to this module is the output from the previous round of the fine-grained completion module. When $u = 0$, $\mathcal{X} = \mathcal{X}_{FI}$, representing that the input for the first round is the initialized $\mathcal{X}$. Based on the research by Xie et al. [26], we refined the processing flow of this module, which mainly consists of the following four steps:

*a.* Grouping: Divide $\mathcal{X}$ into overlapping 3D cubes (spatial dimensions of $W_1 \times W_1$, containing all spectral bands). Each cube is reshaped into a 2D full-band block of size $W_1^2 \times I_3$, forming a set of 2D full-band block collection $\varphi = \{Y_i \in \mathbb{R}^{W^2 \times I_3}\}_{i=1}^{S}$, where $S$ is the number of blocks.

*b.* Clustering: Use the K-means++ algorithm to cluster the 2D full-band blocks, generating L clusters, each containing several non-local similar blocks. The blocks in each cluster are recombined into 3D tensors $\hat{\mathcal{X}}_l \in W_1^2 \times I_3 \times H_l$, where $H_l$ is the number of blocks in the $l$-th cluster.

*c.* Completion: Apply a coarse-grained model based on Tucker decomposition to each clustered 3D tensor $\hat{\mathcal{X}}_l$ for completion:

$$\min_{\mathcal{X}} \sum_{k=1}^{N} \alpha_i R_\varepsilon \left(\hat{\mathcal{X}}_l\right)_{(k)}$$
$$s.t. \ \mathcal{X}_\Omega = \mathcal{T}_\Omega \tag{7}$$

*d.* Reconstruction: Return the completed 3D tensors to their original positions to obtain the reconstructed HSI $\mathcal{X}$, denoted as $\mathcal{X}^{[c,u+1]}$.

**2) Fine-Grained Non-Local Module**

Let the input to this module be the output of the coarse-grained completion module in the current iteration, i.e., $\mathcal{X} = \mathcal{X}^{[c,u+1]}$. Based on the research by Zheng et al. [30], we have organized the processing flow of this module into the following four main steps:

*a.* Grouping: Divide image $\mathcal{X}$ into overlapping patches of size $W_2 \times W_2 \times I_3$, then select key patches at fixed intervals $v$.

*b.* Block matching: For each key block, calculate similarity with other blocks using Euclidean distance, select the $k$ most similar blocks to form an NSS group, then elevate it to a fourth-order tensor (two spatial dimensions, one spectral dimension, and one similarity group dimension).

*c.* Completion: Treat each fourth-order NSS group as an independent unit and complete it using a fine-grained method

based on FCTN decomposition. The optimization problem is as follows:

$$\min_{\mathcal{X},\mathcal{G}} \frac{1}{2} \left\| \hat{\mathcal{X}}_l - FCTN(\mathcal{G}_1, \mathcal{G}_2, \cdots, \mathcal{G}_N) \right\|_F^2$$
$$s.t. \ \mathcal{X}_\Omega = \mathcal{T}_\Omega \qquad (8)$$

*d.* Aggregation: Aggregate all restored NSS groups $\hat{\mathcal{X}}_l$ back to their original positions using a weighted averaging method to obtain the reconstructed image $\mathcal{X}^{[f,u+1]}$ for the current iteration.

*D. Pseudocode*

In Algorithm 1, we summarize the solution process of the proposed multi-grained non-local HSI recovery model.

---

**Algorithm 1** ADMM and PAM Based Optimization Algorithm for MG-NSS Based HSI Recovery Model.

---

**Input:** The observed data $\mathcal{T} \in \mathbb{R}^{I_1 \times I_2 \times I_3}$.

**Parameters:** $W_1, W_2, Iters$.

**Output:** $\mathcal{X}$.
**Initialization:**
  Update $\mathcal{X}_{CI}$ by (4).
  Update $\mathcal{X}_{FI}$ by (6).
$\mathcal{X} = \mathcal{X}_{FI}$.

**While** $u < Iters$ **do:**
  Update $\mathcal{X}^{[c,u+1]}$ by (7).
  Update $\mathcal{X}^{[f,u+1]}$ by (8).
**End for**
$\mathcal{X} = \mathcal{X}^{[f,u+1]}$.

---

IV. EXPERIMENTS

In this section, we will evaluate the recovery performance of the MG-NSS model on multi-spectral and hyper-spectral datasets. Below, we will introduce the experimental setup, experimental results, and ablation analysis in sequence.

*A. Experimental Setup*

1) Datasets

This paper conducted experiments using the multispectral image dataset CAVE and the HSI dataset Pavia University. The simulated degradation patterns include: (1) random pixel missing with sampling rates set at 0.01 and 0.03, and (2) random stripe missing with sampling rates of 0.05 and 0.15.

2) Evaluation Metrics

This study employs Peak Signal-to-Noise Ratio (PSNR) and Structural Similarity Index (SSIM) as quantitative metrics to evaluate recovery performance, where higher values indicate better reconstruction quality. Additionally, computational time is introduced to assess the efficiency of each algorithm.

3) Comparative Algorithms

This work performs extensive comparisons with seven state-of-the-art HSI recovery approaches: High Accuracy Low-Rank Tensor Completion (HaLRTC) [31], Non-Local Similarity and Low-Rank Tensor Completion (NLS-LR) [32], Tensor Train with Overlapping Ket Augmentation (TT-OKA) [33], Fully-Connected Tensor Network decomposition (FCTN) [24], Tensor Wheel decomposition for Tensor Completion (TW-TC) [34], Non-Local Fully-Connected Tensor Network (NL-FCTN) [30], and Tensor Correlated Total Variation (TCTV) [35]. These methods are systematically categorized into coarse-grained tensor decomposition approaches (HaLRTC, NLS-LR, TCTV) and fine-grained tensor decomposition approaches (TT-OKA, FCTN, TW-TC, NL-FCTN) based on their underlying mathematical frameworks and granularity characteristics.

4) Parameter Settings

In the coarse-grained non-local module, the size of similar patch is set to 5, with a stride of 2, parameter $\mu$ to 1/160, $\eta$ to 1.1, and $\alpha_i$ to $[1, 1.5, 1.2]$. For the fine-grained non-local module, the size of similar patch is configured as 6 with a stride of 5.

*B. Experimental Results*

1) CAVE Dataset Experiments

Fig. 2 and Table I present the recovery results on the "Flowers" data from the CAVE dataset (256×256×31). The visual results in Fig. 2 display the reconstructed images at the 30th spectral band, while Table 1 quantitatively compares different methods, with the best performance values highlighted in bold black font for clear identification. Specifically, in the comparative methods, taking the advanced coarse-grained decomposition method TCTV and the advanced fine-grained decomposition method NL-FCTN as examples, in the scenario of random pixel missing, when the sampling rate is 0.01, the observed entries are too sparse. Due to the integration of local smoothness priors by TCTV, its performance outperforms NL-FCTN. However, when the sampling rate is 0.03, NL-FCTN surpasses TCTV, a coarse-grained method that uses local smoothness priors, indicating that fine-grained methods have an advantage in capturing image detail information. In the scenario of random stripe missing, which is a structured missing, the coarse-grained method TCTV demonstrates a prominent advantage, suggesting that coarse-grained methods excel in capturing image structural information. The proposed method MG-NSS in this paper, which integrates tensor coarse-grained decomposition with fine-grained decomposition, achieves the best experimental results in all scenarios, indicating that combining both granularities benefits the simultaneous capture of global structural information and local detail information in images. In terms of time consumption, since MG-NSS requires simultaneous coarse-grained and fine-grained decomposition of HSI, it has a higher time complexity and consumes more time.

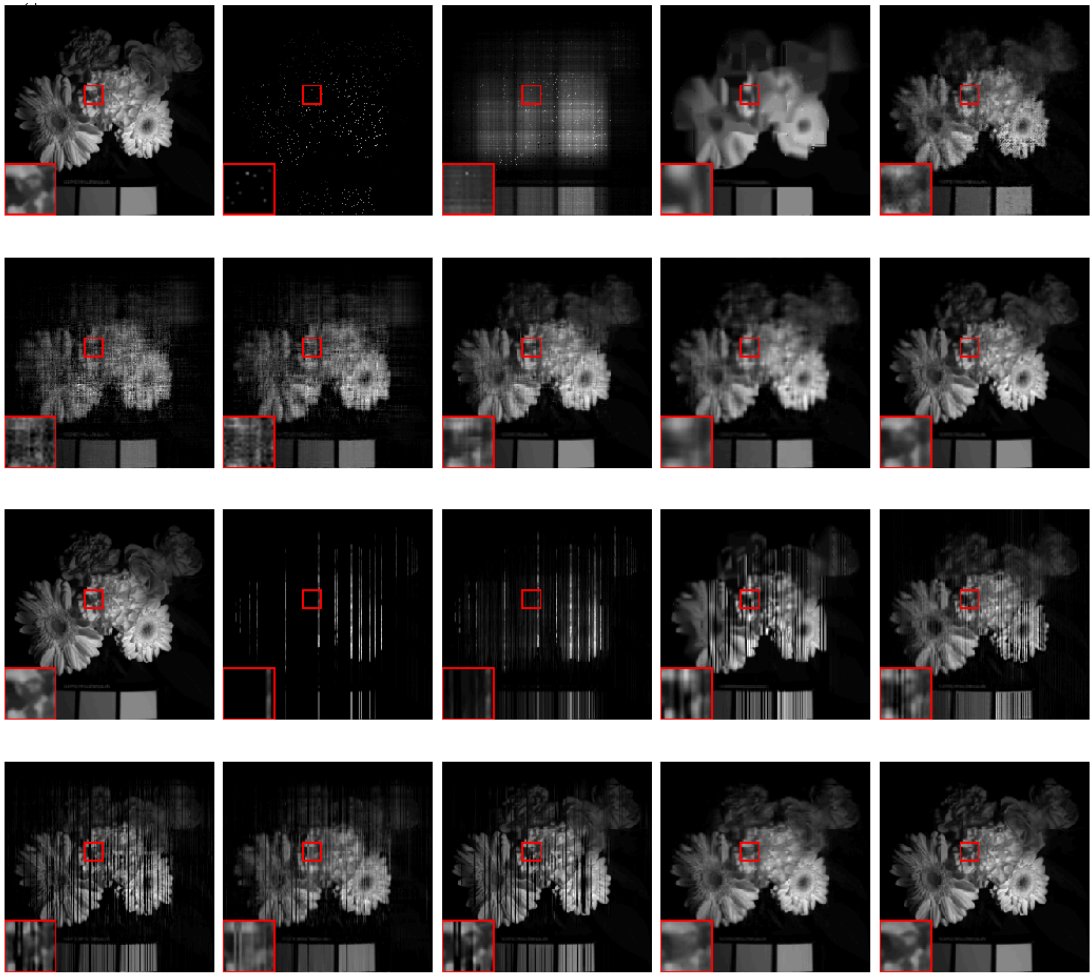

**Fig. 2.** Recovery results on the Flowers dataset. The first and second rows show the recovery results for random pixel missing with a sampling rate of 0.03. The third and fourth rows show the recovery results for random stripe missing with a sampling rate of 0.15.

TABLE I
PSNR AND SSIM METRIC RESULTS ON THE FLOWERS DATASET

| Flowers | SR | Index | HaLRTC | NLS-LR | TT-OKA | FCTN | TW-TC | NL-FCTN | TCTV | MG-NSS |
|---|---|---|---|---|---|---|---|---|---|---|
| Pixels Missing | 0.01 | PSNR | 17.3870 | 26.4421 | 21.8775 | 21.0804 | 22.7624 | 24.1202 | 28.2311 | **31.2265** |
| | | SSIM | 0.5524 | 0.6493 | 0.5571 | 0.4405 | 0.4462 | 0.7071 | 0.7732 | **0.8785** |
| | 0.03 | PSNR | 22.8552 | 29.2351 | 31.6932 | 27.8638 | 28.7754 | 34.4569 | 32.4570 | **37.7814** |
| | | SSIM | 0.6515 | 0.8062 | 0.8324 | 0.6602 | 0.6825 | 0.9307 | 0.8759 | **0.9637** |
| Stripes Missing | 0.05 | PSNR | 17.2779 | 20.4463 | 21.9370 | 18.4917 | 19.8885 | 19.1025 | 26.9916 | **30.6509** |
| | | SSIM | 0.4493 | 0.6521 | 0.4444 | 0.4177 | 0.4108 | 0.5552 | 0.7366 | **0.9116** |
| | 0.15 | PSNR | 20.1907 | 25.2189 | 32.4558 | 26.3722 | 29.5622 | 27.7312 | 37.1584 | **44.2950** |
| | | SSIM | 0.6342 | 0.8322 | 0.8818 | 0.7565 | 0.7574 | 0.8847 | 0.9497 | **0.9904** |
| Average Time (S) | | | 35.47 | 2997.43 | 202.48 | 71.17 | 264.44 | 2255.38 | 190.72 | 3352.23 |

**2) Pavia University Dataset Experiments**

Fig. 3 and Table Ⅱ present the recovery results on the Pavia University dataset. While the size of original HSI is 610×340×103, we cropped a sub-image of 256×256×40 for experimental and analytical convenience. Both visual assessment and quantitative metrics demonstrate that MG-NSS achieves optimal performance across all scenarios, attributable to its integrated coarse-grained and fine-grained tensor decomposition framework. When combined with the experimental analysis on the multispectral CAVE dataset, these results confirm that the proposed MG-NSS model delivers superior recovery performance under diverse scenarios and datasets. This compelling evidence substantiates the method's exceptional capability and strong generalization capacity.

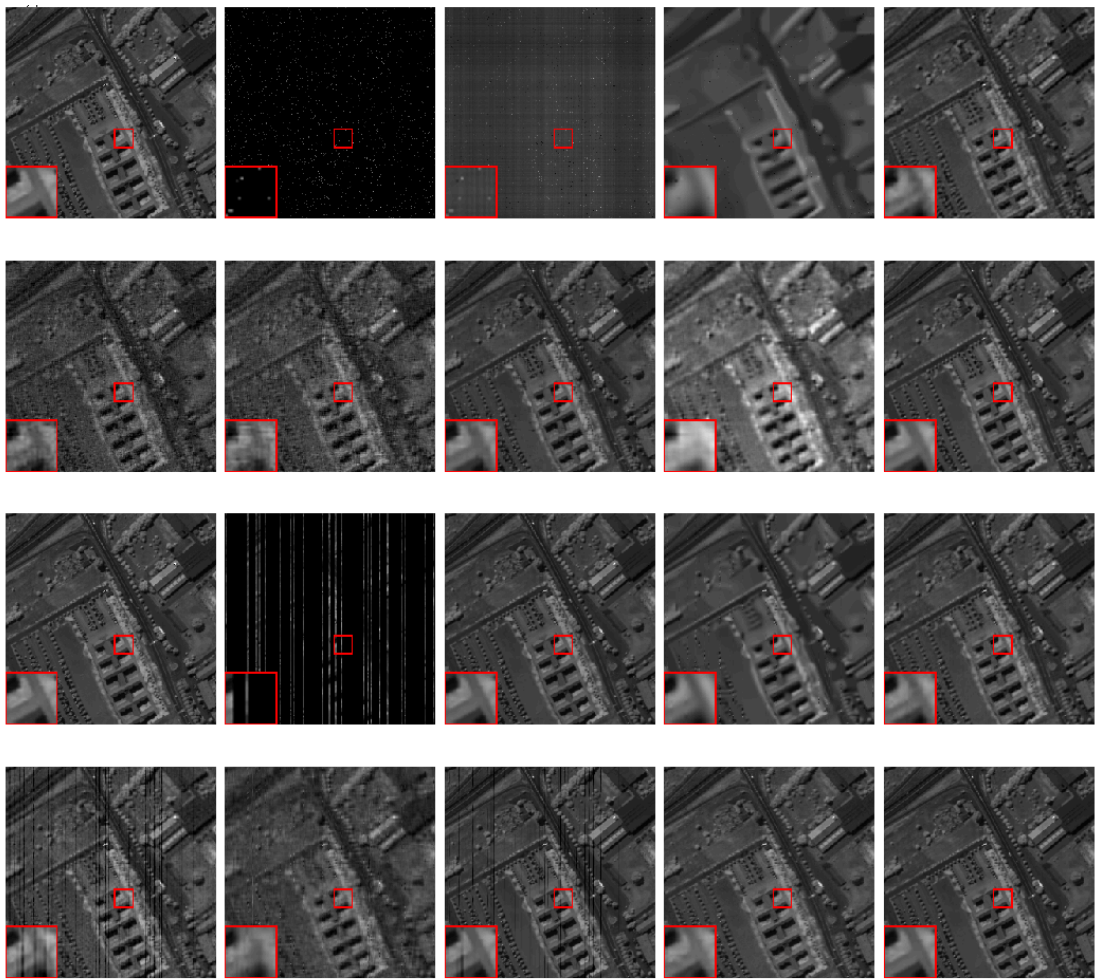

**Fig. 3**. Recovery results on the Pavia University dataset. The first and second rows show the recovery results for random point missing with a sampling rate of 0.03, while the third and fourth rows show the recovery results for random stripe missing with a sampling rate of 0.15.

TABLE II
PSNR AND SSIM METRIC RESULTS ON THE PAVIA UNIVERSITY DATASET

| PaviaU | SR | Index | HaLRTC | NLS-LR | TT-OKA | FCTN | TW-TC | NL-FCTN | TCTV | MG-NSS |
|---|---|---|---|---|---|---|---|---|---|---|
| Pixels Missing | 0.01 | PSNR | 19.0399 | 22.7252 | 25.7286 | 18.3709 | 19.4836 | 21.7688 | 23.6293 | **27.3420** |
| | | SSIM | 0.3526 | 0.4738 | 0.6603 | 0.1576 | 0.1935 | 0.4439 | 0.5408 | **0.7621** |
| | 0.03 | PSNR | 20.3841 | 25.2440 | 29.7332 | 25.2201 | 24.7420 | 31.0606 | 27.0783 | **33.9947** |
| | | SSIM | 0.3982 | 0.6012 | 0.8298 | 0.5790 | 0.5410 | 0.8828 | 0.7324 | **0.9397** |
| Stripes Missing | 0.05 | PSNR | 13.1212 | 17.3049 | 29.8538 | 14.4447 | 21.4263 | 15.0035 | 24.2121 | **32.6421** |
| | | SSIM | 0.0815 | 0.1830 | 0.8388 | 0.1279 | 0.3853 | 0.1704 | 0.6654 | **0.9225** |
| | 0.15 | PSNR | 19.0731 | 29.5516 | 32.1123 | 24.9529 | 26.5741 | 30.4069 | 35.1710 | **41.1306** |
| | | SSIM | 0.3714 | 0.8220 | 0.9008 | 0.6624 | 0.6691 | 0.8856 | 0.9496 | **0.9866** |
| Average Time (S) | | | 46.52 | 3870.56 | 258.41 | 79.29 | 315.08 | 2782.63 | 231.30 | 4404.06 |

*C. Ablation Analysis*

The comparative experiments with various HSI recovery methods have already demonstrated the superior performance of the MG-NSS method proposed in this paper. To further analyze the impact of the coarse-grained component and the fine-grained component of MG-NSS on the overall performance of the model, we conducted ablation experiments on the Flowers dataset. Table III shows the ablation results for point dropout with a sampling rate of 0.01 and stripe dropout with a sampling rate of 0.05, where C represents the coarse-grained component, and F represents the fine-grained component, including their respective initialization and non-local processing processes. The data in the table indicates that the coarse-grained and fine-grained components play different roles, and their combined use achieves the best results.

TABLE III
ABLATION ANALYSIS RESULTS ON THE FLOWERS DATASET

| Scenes | Index | C √ | F × | C × | F √ | C √ | F √ |
|---|---|---|---|---|---|---|---|
| Pixels Missing | PSNR | 29.4199 | | 28.1237 | | **31.2265** | |
| | SSIM | 0.7389 | | 0.7806 | | **0.8785** | |
| Stripes Missing | PSNR | 24.9369 | | 27.5817 | | **30.6509** | |
| | SSIM | 0.8204 | | 0.7714 | | **0.9116** | |

## V. CONCLUSION

This paper proposes a multi-grained non-local model for HSI recovery. Under the non-local framework, the model integrates tensor coarse-grained decomposition and fine-grained decomposition, respectively for capturing the global structural information and local detail information of HSI. Extensive experiments demonstrate the effectiveness and superiority of this method.

In future work, we will continue to conduct in-depth research on the granularity of tensor decomposition, providing error analysis and boundary conditions for missing data completion. Building on this, we will integrate deep priors to further improve the recovery accuracy of HSI.